\documentclass[letterpaper, 10 pt, conference]{ieeeconf}

\usepackage{amsmath,amsfonts}
\usepackage{algorithmic}
\usepackage{algorithm}
\usepackage{array}
\usepackage[caption=false,font=normalsize,labelfont=sf,textfont=sf]{subfig}
\usepackage{textcomp}
\usepackage{stfloats}
\usepackage{url}
\usepackage{graphicx}
\usepackage{cite}
\usepackage{booktabs}
\usepackage{multirow}
\usepackage{makecell}
\usepackage{hyperref}
\usepackage{xcolor}
\usepackage[table]{xcolor}

\IEEEoverridecommandlockouts
\title{\LARGE \bf
PlanRL: A Trajectory \underline{Plan}ning Architecture for \\
\underline{R}einforcement \underline{L}earning-based Driving Experts
}

\author{
Joonhee Lim$^{1,\dagger}$,
Yongjae Lee$^{2,\dagger}$,
Jangho Shin$^{3}$,
and Dongsuk Kum$^{2}$%
\thanks{$^{\dagger}$J. Lim and Y. Lee contributed equally to this work.}
\thanks{This work was supported by Hyundai Motor Company and the National Research Foundation of Korea (NRF) grant funded by the Korea government (MSIT) (RS-2026-25479609).}
\thanks{$^{1}$J. Lim is with the Robotics Program, Korea Advanced Institute of Science and Technology (KAIST), Daejeon 34141, South Korea (email: {\tt\small kingear3@kaist.ac.kr})}
\thanks{$^{2}$Y. Lee and D. Kum are with the Cho Chun Shik Graduate School of Mobility, Korea Advanced Institute of Science and Technology (KAIST), Daejeon 34141, South Korea (email: {\tt\small yongjae.lee@kaist.ac.kr}, {\tt\small dskum@kaist.ac.kr})}
\thanks{$^{3}$J. Shin is with Hyundai Motor Company, Hwaseong 18280, South Korea (email: {\tt\small jangho\_shin@hyundai.com})}
}

\begin{document}

\maketitle
\thispagestyle{empty}
\pagestyle{empty}

\begin{abstract}

Reinforcement learning (RL) has become a prominent framework for developing driving experts in autonomous vehicles. However, most existing RL-based experts are designed to output direct control commands (e.g., throttle, steering), which suffer from a lack of interpretability, high spatial complexity in learning road geometries, and poor compatibility with modern end-to-end planning architectures. To address these limitations, we propose a novel trajectory planning architecture for RL driving experts that integrates an RL policy with a polynomial-based trajectory planner. By employing a Frenet-frame coordinate system, our method simplifies complex road geometries into a curvilinear framework, offering a structured coordinate prior that facilitates policy learning. Furthermore, we incorporate a kinematic feasibility check into the planning stage to ensure that generated trajectories remain within the vehicle's physical limits, effectively mitigating cumulative tracking errors typically found in planning-based systems. We evaluate our approach on key CARLA benchmarks, where it significantly outperforms existing state-of-the-art control-based RL experts. On the CARLA Offline Leaderboard v1 and NoCrash benchmarks, our method improves the driving score by 5\% and 11\%, respectively, and increases the success rate by 8\% and 19\%.

\end{abstract}
\section{INTRODUCTION}

Reinforcement learning (RL) has emerged as a powerful framework for developing driving experts, addressing the limitations of rule-based experts~\cite{roach}. Unlike rule-based methods that require laborious manual tuning and lack scalability, RL-based experts progressively accumulate driving knowledge by exploring diverse and challenging scenarios within simulated environments~\cite{rlgood}. Beyond providing raw actions, these models generate rich latent features that offer more informative guidance than rule-based systems. This allows RL-based experts to provide a more effective foundation for autonomous driving compared to traditional methods.

Most existing studies in this field are structured to output direct control commands, such as throttle, steering, and brake~\cite{roach,carl,driveadapter,think2drive,raw2drive,adawm}. This architectural preference exists because methods that plan and follow trajectories often suffer from cumulative tracking errors during execution, which significantly degrade performance on high-curvature roads~\cite{tcp}. Furthermore, while real-world applications can leverage model predictive control (MPC) based on vehicle dynamics~\cite{mpc}, simulators like CARLA~\cite{carla} lack the precise dynamics required to make such model-based controllers effective. Consequently, direct control has been favored as a pragmatic solution to ensure immediate responsiveness within simulated environments.

\begin{figure}[!t]
\centering
\includegraphics[width=1.0\columnwidth]{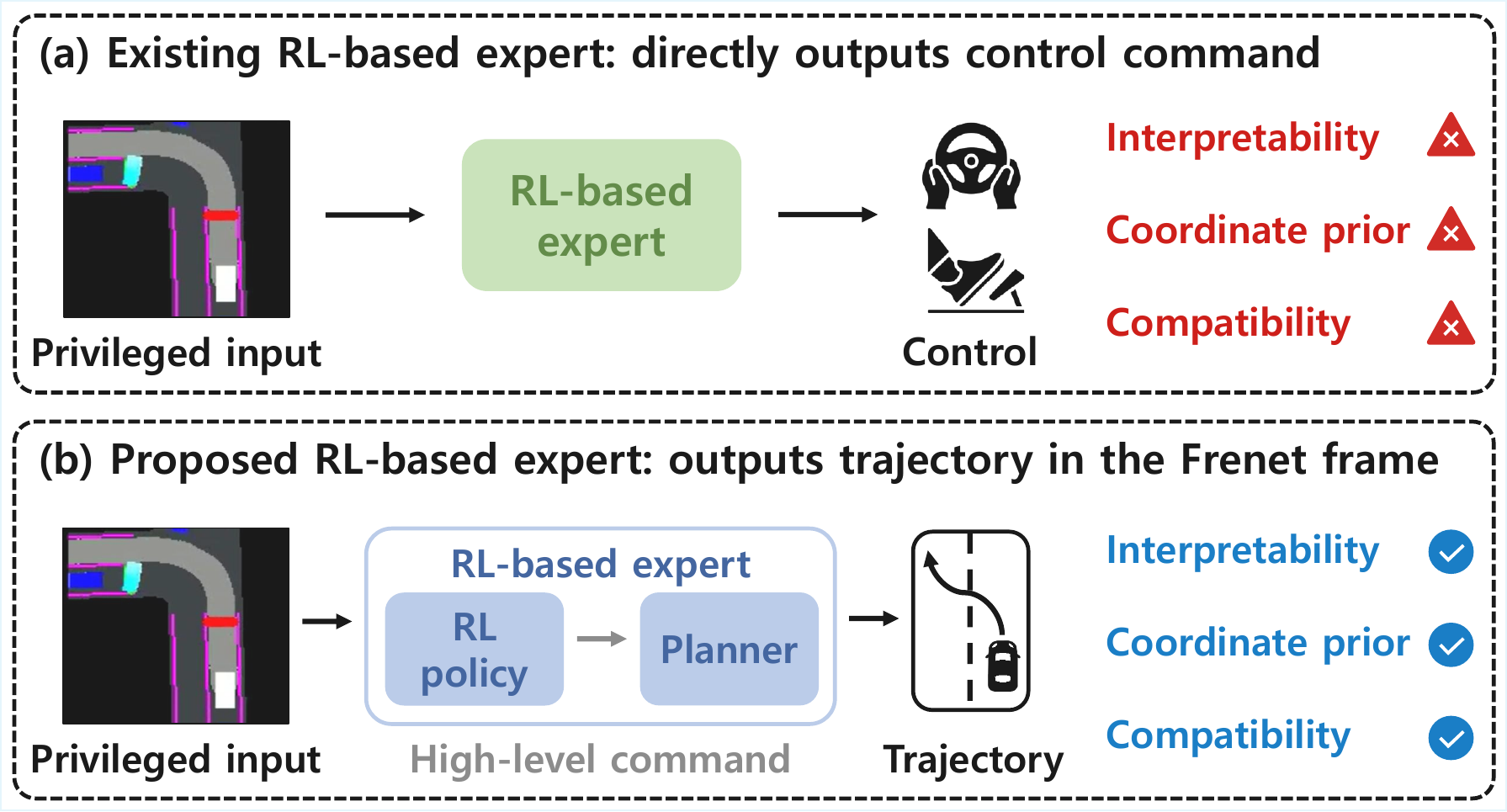}
\caption{Comparison of the proposed RL expert with an existing RL expert. While conventional RL-based experts directly output control commands, the proposed method employs an RL policy to output high-level commands, which a planner uses to generate trajectories.}
\vspace{-12pt}
\label{introduction}
\end{figure}

However, such control-centric RL experts face significant limitations that can be summarized into three primary challenges. First, they lack interpretability; direct control commands provide no insight into the vehicle's intended path, making it difficult to validate the decision-making process~\cite{lim}. Second, they place a heavy burden on the model to internalize diverse and complex road geometries without the aid of a structured coordinate framework. This forces the model to learn spatial mappings for every variation in curvature across diverse urban environments, often resulting in sub-optimal policies. Third, the reliance on raw control values severely limits their utility as experts for real-world end-to-end (E2E) autonomous driving systems. Contemporary state-of-the-art models~\cite{distilldrive,drivesuprim} on real-world planning benchmarks, such as nuScenes~\cite{nuscenes} and NAVSIM~\cite{navsim}, predominantly adopt trajectory planning as their core architecture. Consequently, an expert model confined to control commands remains decoupled from these established architectural standards, failing to provide the structured supervision required to effectively guide the learning of modern E2E planning pipelines\cite{uniad}.

To address these challenges, this study proposes a novel trajectory planning architecture for RL-based experts utilizing privileged information, as shown in Fig.~\ref{introduction}. By operating within the Frenet-frame coordinate system, our architecture effectively simplifies the representation of complex road geometries, ensuring robust performance even in high-curvature urban scenarios. Furthermore, the proposed framework generates trajectories through feasible longitudinal and lateral motions, ensuring the planned paths remain within the vehicle's physical limits. By integrating this kinematic feasibility directly into the planning stage, the architecture minimizes the cumulative tracking errors typically introduced by the addition of a planner, thereby bridging the gap between high-level planning and low-level execution.

We evaluate the proposed method on key CARLA benchmarks, including the Offline Leaderboard v1~\cite{carla_leaderboard} and NoCrash~\cite{nocrash}, where it significantly outperforms existing state-of-the-art control-based RL experts. Specifically, our approach improves the driving score by 5\% and 11\%, and increases the success rate by 8\% and 19\% on the Offline Leaderboard v1~\cite{carla_leaderboard} and NoCrash~\cite{nocrash}, respectively. The key contributions of this paper are summarized below:
\begin{itemize}
    \item To the best of our knowledge, this study is the first to propose a trajectory planning architecture for RL-based experts within the CARLA benchmark.
    \item A robust hybrid planning framework is formulated by seamlessly integrating the Frenet-frame coordinate system with a kinematic feasibility check. By projecting complex road geometries into a curvilinear framework, the spatial complexity of the environment is reduced, allowing for more stable planning.
    \item The superiority of the proposed architecture is demonstrated through extensive evaluations on key benchmarks. It is shown that the proposed method significantly outperforms existing state-of-the-art control-based RL experts.
\end{itemize}
\section{RELATED WORK}

\subsection{Rule-based Experts for Autonomous Driving}

Rule-based experts operate based on predefined decision logic and handcrafted rules, often utilizing privileged states to generate more accurate and reliable driving behaviors. They played a foundational role in early autonomous driving research. A primary example of such an expert is the CARLA Autopilot, which leverages direct access to ground-truth simulation states to achieve high performance, particularly on the CARLA Leaderboard v1~\cite{carla_leaderboard}. It follows handcrafted rules for surrounding vehicles, pedestrians, and traffic lights, while performing route-based planning using privileged map and state information. Recently, PDM-Lite~\cite{pdmlite} has emerged as an expert for the CARLA Leaderboard v2~\cite{carla_leaderboard}, successfully navigating all benchmark scenarios in this version. It achieves this by identifying scenario types and applying specialized rules tailored to each case. SEED~\cite{seed}, the Transfuser variants~\cite{transfuser,transfuser++}, CILRS~\cite{nocrash}, and NEAT~\cite{neat} also develop rule-based experts designed for specific benchmarks, such as Town05 and Longest6~\cite{transfuser}. However, these systems face fundamental limitations in scalability and require extensive manual tuning; for instance, PDM-Lite~\cite{pdmlite} adopts different hyperparameters for different scenarios, demanding heavy manual labor~\cite{think2drive}. Furthermore, unlike learning-based experts that provide rich latent features beyond raw control commands, rule-based systems typically offer only explicit actions, limiting their ability to supply informative intermediate representations for downstream learning. As an alternative, approaches such as WOR~\cite{wor} and LBC~\cite{lbc} leverage rule-based experts to train learning-based experts. However, these methods essentially remain heavily dependent on the performance of the underlying heuristic rule-based experts.

\subsection{RL-based Experts for Autonomous Driving}

RL has emerged as a promising paradigm for autonomous driving experts, enabling policies to be learned directly from interaction with the environment and to adapt to complex, dynamic scenarios without relying on handcrafted rules.

Among such RL-based experts, Roach~\cite{roach} establishes a strong foundation by demonstrating that RL can serve as an effective expert for imitation learning without requiring human demonstration datasets. Building upon Roach~\cite{roach}, DriveAdapter~\cite{driveadapter} combines Roach with emergency brake rules to further stabilize decision-making and significantly improve overall performance. CaRL~\cite{carl} also builds upon Roach~\cite{roach} and revisits its reward design, identifying the limitations of heavily shaped rewards commonly used in RL for autonomous driving. Instead of relying on complex rewards, CaRL proposes a simplified objective that primarily maximizes route completion while applying multiplicative penalties for infractions. This streamlined formulation enables RL to scale effectively to larger batch sizes and improves training stability over extensive training steps.

In parallel, a line of research introduces model-based RL frameworks that incorporate learned world models to enhance data efficiency and planning capability. Think2Drive~\cite{think2drive} pioneers this direction in CARLA Leaderboard v2 by learning a latent world model that predicts environment transitions and serves as a neural simulator for training the planner. By operating in a low-dimensional latent space with parallel tensor computation, it significantly improves training efficiency while achieving strong performance on CARLA Leaderboard v2. Subsequent works further refine this paradigm: Raw2Drive~\cite{raw2drive} extends the RL expert paradigm by transferring privileged world model knowledge to a raw sensor-based model, while AdaWM~\cite{adawm} proposes an adaptive world model to mitigate dynamics and policy mismatches in world model-based planning (their official source code is not publicly available, and thus we are unable to include them in our experimental comparison).

Despite these advances, existing RL-based approaches still inherit fundamental limitations from their control-centric architectures. By directly outputting low-level control commands, they lack interpretability and require the policy to implicitly learn complex and diverse road geometries without explicit structural representations or coordinate priors. Moreover, their reliance on control outputs reduces their effectiveness as experts for real-world E2E systems, where leading benchmarks such as nuScenes~\cite{nuscenes} and NAVSIM~\cite{navsim} predominantly adopt trajectory planning architectures. Consequently, they fall short of delivering the structured supervision required by modern E2E planning pipelines.
\begin{figure*}[!t]
\centering
\includegraphics[width=0.95\textwidth]{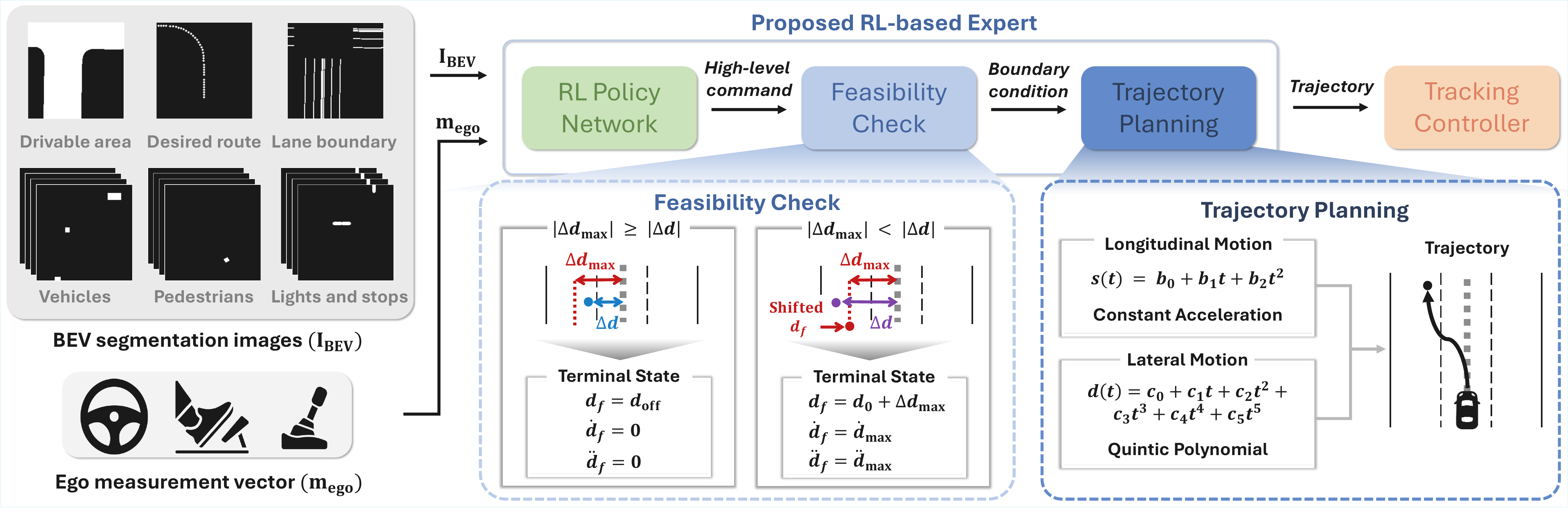}
\vspace{-8pt}
\caption{\textbf{Overall architecture of the proposed RL-based expert: PlanRL.}  The RL Policy Network Module outputs a high-level command using BEV segmentation images and an ego measurement vector as inputs. The Feasibility Check Module adjusts the terminal lateral state to satisfy kinematic constraints. The Trajectory Planning Module then generates a smooth trajectory based on the adjusted terminal state.}
\vspace{-10pt}
\label{framework}
\end{figure*}

\section{METHODOLOGY}

\subsection{Overview}
This study proposes a novel trajectory planning architecture for RL-based experts in autonomous driving. The proposed method consists of three main components: an RL policy, a kinematic feasibility check, and a trajectory planning module (Fig.~\ref{framework}). The RL policy outputs high-level commands, which are converted into kinematically feasible trajectories through the feasibility check and trajectory planning modules, and then tracked by a PID controller.

\subsection{Reinforcement Learning Policy}
We adopt the architecture of Roach~\cite{roach}, an RL-based driving expert trained via Proximal Policy Optimization (PPO)~\cite{ppo}, as an RL policy to output high-level commands for trajectory planning.
The key components of the RL policy are described below: input, output representation, reward function, network architecture, and training objective.

\textbf{Input Representation.}
The proposed method takes as input a Bird’s-Eye View (BEV) semantic segmentation image $\mathbf{I}_\text{BEV} \in [0, 1]^{W \times H \times C}$ with $W = 192$ px, $H = 192$ px, and $C = 15$ channels at 5 px/m resolution, encoding: 
\begin{itemize}
    \item \textit{Channels 1-2}: Drivable area and desired route (While Roach~\cite{roach} represents routes as polygons, the proposed method represents them as lines, which has been experimentally shown to achieve higher performance.)
    \item \textit{Channel 3}: Lane boundary (solid: white, dashed: gray)
    \item \textit{Channels 4-7}: Vehicle bounding boxes at $t \in \{-1.5, -1.0, -0.5, 0\}$ s
    \item \textit{Channels 8-11}: Pedestrian bounding boxes at $t \in \{-1.5, -1.0, -0.5, 0\}$ s, rendered at twice actual size
    \item \textit{Channels 12-15}: Traffic lights and stop signs, with brightness encoding the state (same temporal sequence as vehicles and pedestrians)
\end{itemize}
In addition to the BEV image, it takes a 6D measurement vector  $\mathbf{m}_\text{ego} \in \mathbb{R}^6$ as input to capture ego-vehicle states not represented in the visual observation: steering, throttle, brake, gear, and longitudinal and lateral speeds.

\textbf{Output Representation.}
The RL policy outputs high-level commands as actions, which distinguishes it from Roach~\cite{roach}, which outputs direct control commands. The high-level commands include a two-dimensional action vector $\mathbf{a} = [a_\text{lon}, a_\text{lat}]^\top$, where $a_\text{lon}$ represents the longitudinal acceleration command and $a_\text{lat}$ represents the target lateral offset from the reference path. They are then passed through a feasibility check and subsequently utilized by the trajectory planning module to generate the final trajectory.

We employ a Beta distribution $\mathcal{B}(\alpha, \beta)$ with shape parameters $\alpha, \beta > 0$ to parameterize continuous actions. This distribution naturally produces action outputs bounded within the $(0, 1)$ interval, which are then linearly rescaled to the ranges used for high-level commands: longitudinal acceleration $a_\text{lon} \in [-3.2, 3.5]$ m/s$^2$ and target lateral offset $a_\text{lat} \in [-3.5, 3.5]$ m. The acceleration limits are selected to ensure a balance between motion comfort and operational maneuverability in urban environments~\cite{iso_limit}. The lateral bound of $\pm 3.5$ m is selected to encompass standard lane widths, enabling both precise lane-keeping and complete lane-change maneuvers within a single planning cycle.

Unlike Gaussian distributions, which suffer from estimation bias in bounded action spaces due to boundary effects, the Beta distribution is inherently bounded and bias-free~\cite{beta}. This eliminates the need for boundary clipping or squashing transformations such as tanh, thereby avoiding the gradient distortion typically introduced by such auxiliary layers. Furthermore, reduced gradient variance leads to more stable and consistent optimization. The Beta distribution also enables explicit computation of key metrics including entropy and KL-divergence, leading to more stable and efficient policy updates. This stability is particularly valuable in complex driving scenarios, where the agent must maintain high-precision control while exploring diverse strategies near the physical limits of the action space.

\textbf{Reward Function.}
The RL policy is trained using a dense reward signal that encourages safe and smooth driving behavior. The total reward at timestep $t$ is expressed as the sum of five components:
\begin{align}
R_t &= R_{\text{spd},t} + R_{\text{pos},t} + R_{\text{rot},t} + R_{\text{act},t} + R_{\text{term},t}.
\end{align}

The speed reward $R_{\text{spd},t}$ is defined as:
\begin{align}
R_{\text{spd},t} &= 1 - \frac{|v_t - v_t^\star|}{v_{\max}},
\end{align}
where $v_t$ is the current speed, $v_t^\star$ is the desired speed dynamically adjusted based on nearby vehicles, pedestrians, traffic lights, and active stop signs, and $v_{\max} = 6$ m/s. The desired speed computation follows Roach~\cite{roach}.

The position reward $R_{\text{pos},t}$ is defined as:
\begin{align}
R_{\text{pos},t} &= -\frac{1}{2} |e_{\perp,t}|,
\end{align}
where $e_{\perp,t}$ is the lateral deviation from the route centerline, measured in meters.

The rotation reward $R_{\text{rot},t}$ is defined as:
\begin{align}
R_{\text{rot},t} &= -|\Delta\psi_t|,
\end{align}
where $\Delta\psi_t = \psi_{\text{ego},t} - \psi_{\text{route},t}$ is the heading error (in radians, cast to $[-\pi, \pi]$).

The action reward $R_{\text{act},t}$ is defined as:
\begin{align}
R_{\text{act},t} &=
\begin{cases}
-0.1, & \text{if } |\delta_t - \delta_{t-1}| > 0.01, \\
0,    & \text{otherwise},
\end{cases}
\end{align}
where $\delta_t$ is the steering command, and this term discourages abrupt steering changes.

The terminal reward $R_{\text{term},t}$ is defined as:
\begin{align}
R_{\text{term},t} &=
\begin{cases}
-1 - v_t, & \text{collision or traffic infraction},\\
-1,       & \text{route deviation/blockage},\\
0,        & \text{otherwise},
\end{cases}
\end{align}
where $R_{\text{term},t}$ applies penalties for collisions, traffic infractions, route deviations, or blockages. A route deviation is triggered when the lateral distance to the desired route centerline exceeds 3.5 m, and a blockage is triggered when the ego vehicle's speed remains below 0.1 m/s for over 90 seconds. This reward is issued upon episode termination.

\textbf{Network Architecture.}
The policy network $\pi_\theta$ and value network $V_\phi$ share the same feature extractors, which consist of a BEV encoder (6 convolutional layers extracting 256-dimensional features) and a measurement encoder (2 fully-connected layers). The policy head comprises 2 FC layers with 256 hidden units each, outputting Beta distribution parameters $(\alpha, \beta) \in \mathbb{R}^2_+$ for each action dimension. The value head uses a separate branch of 2 FC layers with 256 units, outputting a scalar estimate of the expected return.

\textbf{Training Objective.}
The policy network is trained using PPO augmented with entropy regularization and event-conditioned exploration priors proposed in Roach~\cite{roach}. The value network is trained concurrently by minimizing the mean squared error between predicted state values and target returns. Network parameters $\theta$ (policy) and $\phi$ (value) are updated using a single Adam optimizer~\cite{adam} with a learning rate of $10^{-5}$. The total loss is expressed as follows:
\begin{align}
\mathcal{L}_{\text{total}} &= \scalebox{0.90}{$
\mathcal{L}_{\text{ppo}}(\theta) 
+ w_{\text{ent}} \mathcal{L}_{\text{ent}}(\theta) 
+ w_{\text{exp}} \mathcal{L}_{\text{exp}}(\theta) 
+ w_{\text{val}} \mathcal{L}_{\text{val}}(\phi)
$},
\end{align}
where $w_{\text{ent}}$, $w_{\text{exp}}$, and $w_{\text{val}}$ are the respective weights for the entropy, event-conditioned exploration, and value losses.

The PPO loss $\mathcal{L}_{\text{ppo}}$ is defined as follows:
\begin{align}
\mathcal{L}_{\text{ppo}}(\theta) &= \scalebox{0.90}{$
-\mathbb{E} \Big[ \min \big( r_t(\theta)\hat{A}_t, \text{clip}(r_t(\theta), 1-\epsilon, 1+\epsilon)\hat{A}_t \big) \Big]
$},
\end{align}
where $r_t(\theta) = \frac{\pi_\theta(a_t|s_t)}{\pi_{\theta_\text{old}}(a_t|s_t)}$ is the ratio of probabilities under the current and old policies, $\hat{A}_t$ is the advantage estimate, and $\text{clip}(x, a, b)$ clamps $x$ to $[a, b]$.

The entropy loss, which encourages the action distribution toward a uniform prior, is defined as follows:
\begin{align}
\mathcal{L}_{\text{ent}}(\theta) &= - \mathbb{E} \big[ H(\pi_\theta(\cdot|s_t)) \big],
\end{align}
where $H(\cdot)$ denotes the entropy of the action distribution and $s_t$ denotes the state at timestep $t$.

The event-conditioned exploration loss, like the entropy loss, encourages exploration and is defined as:
\begin{align}
\mathcal{L}_{\text{exp}}(\theta) &= \scalebox{0.90}{$
\mathbb{E} \Big[ \mathbb{I}_{\{T_{term}-N_z+1, \dots, T_{term}\}}(t) \, \text{KL} (\pi_\theta(\cdot|s_t) \| p_z) \Big]
$},
\end{align}
where $\mathbb{I}_{\{\cdot\}}(t)$ is an indicator function that activates over the final $N_z=100$ timesteps before episode termination at timestep $T_{term}$, $\text{KL}(\cdot \| \cdot)$ denotes the Kullback-Leibler divergence, and $p_z$ is the event-specific prior distribution. For collision events or traffic light/sign violations, we apply a longitudinal action prior $p_z = \mathcal{B}(1, 2.5)$ to encourage deceleration, while leaving the lateral action distribution unchanged. 
In blockage scenarios, a longitudinal prior $\mathcal{B}(2.5, 1)$ is used to promote forward motion. 
For route deviations, a uniform prior $\mathcal{B}(1, 1)$ is imposed on the lateral action.

The value loss is defined as follows:
\begin{align}
\mathcal{L}_{\text{val}}(\phi) &= \mathbb{E} \big[ (V_\phi(s_t) - \hat{R}_t)^2 \big],
\end{align}
where $\hat{R}_t$ is the target return. Both $\hat{A}_t$ and $\hat{R}_t$ are computed using Generalized Advantage Estimation (GAE)~\cite{gae} with discount factor $\gamma=0.99$ and GAE parameter $\lambda=0.9$.

\subsection{Trajectory Planning in the Frenet Frame}

Trajectory generation is performed in the Frenet-frame coordinate system, which provides a convenient curvilinear representation of the road and simplifies planning along complex road geometries.

\textbf{Reference Path Construction.}
A reference path is first constructed from route waypoints used to create the BEV observation. The waypoints are interpolated and smoothed using a Cubic Spline to generate a continuous reference line $\mathbf{r}(s) = [r_x(s),\, r_y(s)]^\top$ parameterized by arc length $s$, where $\tau \in [0, u]$ is the curve parameter and
\begin{equation} s = \int_0^u \sqrt{\left(\frac{dr_x}{d\tau}\right)^2 + \left(\frac{dr_y}{d\tau}\right)^2} d\tau.
\end{equation}

\textbf{Frenet Coordinate Projection.}
The ego vehicle's current state is projected onto the reference line to obtain Frenet initial conditions: longitudinal position $s_0$, longitudinal velocity $\dot{s}_0$, and acceleration $\ddot{s}_0$; lateral offset $d_0$, lateral velocity $\dot{d}_0$, and acceleration $\ddot{d}_0$. 
This projection rotates the global position $(p_x^\text{g}, p_y^\text{g})$ relative to the nearest reference point $(r_x^*, r_y^*)$, as well as the global velocity $(v_x^\text{g}, v_y^\text{g})$ and acceleration $(a_x^\text{g}, a_y^\text{g})$, into the Frenet frame using the reference heading $\psi = \arctan2(dr_y/ds, dr_x/ds)$ at the nearest point:
\begin{align}
\begin{bmatrix} s_0 \\ d_0 \end{bmatrix}
&= \begin{bmatrix} \cos\psi & \sin\psi \\ -\sin\psi & \cos\psi \end{bmatrix}
\begin{bmatrix} p_x^\text{g} - r_x^* \\ p_y^\text{g} - r_y^* \end{bmatrix}, \\
\begin{bmatrix} \dot{s}_0 \\ \dot{d}_0 \end{bmatrix}
&= \begin{bmatrix} \cos\psi & \sin\psi \\ -\sin\psi & \cos\psi \end{bmatrix}
\begin{bmatrix} v_x^\text{g} \\ v_y^\text{g} \end{bmatrix}, \\
\begin{bmatrix} \ddot{s}_0 \\ \ddot{d}_0 \end{bmatrix}
&= \begin{bmatrix} \cos\psi & \sin\psi \\ -\sin\psi & \cos\psi \end{bmatrix}
\begin{bmatrix} a_x^\text{g} \\ a_y^\text{g} \end{bmatrix}.
\end{align}

\textbf{Longitudinal Motion Model.}
The Frenet-frame formulation decouples the two-dimensional motion planning problem into
longitudinal and lateral subproblems. The longitudinal motion is modeled as a second-order
polynomial under a constant-acceleration assumption:
\begin{equation}
s(t) = b_0 + b_1 t + b_2 t^2.
\end{equation}

Following the polynomial formulation, the coefficient vector $[b_0,b_1,b_2]^\top$ is obtained as
\begin{equation}
\begin{bmatrix}
b_0 \\ b_1 \\ b_2
\end{bmatrix}
=
\begin{bmatrix}
1 & 0 & 0 \\
0 & 1 & 0 \\
0 & 0 & 2
\end{bmatrix}^{-1}
\begin{bmatrix}
s(0) \\ \dot{s}(0) \\ \ddot{s}(0)
\end{bmatrix}.
\end{equation}

The boundary conditions are given by
\begin{equation}
s(0)=s_0,\qquad \dot{s}(0)=\dot{s}_0,\qquad \ddot{s}(0)=a_\text{lon},
\end{equation}
where $a_\text{lon}$ denotes the longitudinal acceleration by the RL policy.
To prevent reversing along the reference path, non-negativity constraints are applied to position and velocity:
\begin{equation}
s(t)=\max(s(t),0),\qquad \dot{s}(t)=\max(\dot{s}(t),0).
\end{equation}
These quantities define the speed profile along the trajectory.

\textbf{Lateral Motion Model.}
The lateral motion is planned as a quintic (5th-order) polynomial connecting the initial lateral state $(d_0, \dot{d}_0, \ddot{d}_0)$ to the terminal lateral state
$(d_f, \dot{d}_f, \ddot{d}_f)$ over the planning horizon $T = 1.0$ s:
\begin{equation}
d(t) = c_0 + c_1 t + c_2 t^2 + c_3 t^3 + c_4 t^4 + c_5 t^5.
\end{equation}
The six coefficients $[c_0, c_1, c_2, c_3, c_4, c_5]^\top$ are uniquely determined by the six boundary conditions: position, velocity, and acceleration at both $t = 0$ and $t = T$. Following the standard quintic polynomial formulation, these coefficients are computed as:
\begin{equation}\label{eq:polynomial_coeffs}
\resizebox{0.85\columnwidth}{!}{$
\begin{bmatrix}
c_0 \\ c_1 \\ c_2 \\ c_3 \\ c_4 \\ c_5
\end{bmatrix}
=
\begin{bmatrix}
1 & 0 & 0 & 0 & 0 & 0 \\
0 & 1 & 0 & 0 & 0 & 0 \\
0 & 0 & 2 & 0 & 0 & 0 \\
1 & T & T^2 & T^3 & T^4 & T^5 \\
0 & 1 & 2T & 3T^2 & 4T^3 & 5T^4 \\
0 & 0 & 2 & 6T & 12T^2 & 20T^3
\end{bmatrix}^{-1}
\begin{bmatrix}
d(0) \\ \dot{d}(0) \\ \ddot{d}(0) \\ d(T) \\ \dot{d}(T) \\ \ddot{d}(T)
\end{bmatrix}.
$}
\end{equation}
The boundary conditions are given by
\begin{align}
    d(0) &= d_0,   & \dot{d}(0) &= \dot{d}_0, & \ddot{d}(0) &= \ddot{d}_0, \label{eq:initial_states} \\
    d(T) &= d_f,   & \dot{d}(T) &= \dot{d}_f, & \ddot{d}(T) &= \ddot{d}_f, \label{eq:terminal_states}
\end{align}
where $d_f,\ \dot d_f,\ \ddot d_f$ denote the terminal conditions determined by the feasibility check in the following section.

\subsection{Kinematic Feasibility Check}

This module ensures kinematic feasibility of the RL-generated lateral command.
Subsequently, the coefficients of the quintic polynomial are determined, and the final trajectory is generated.
Given the commanded offset $d_{\text{off}} = a_{\text{lat}}$ and the required displacement $\Delta d = d_{\text{off}} - d_0$, the maximum achievable lateral displacement within the horizon $T$ under $|\ddot{d}| \leq |\ddot{d}_{\text{max}}|$ is
\begin{equation}
    \Delta d_\text{max} = \dot{d}_0\, T
    + \frac{1}{2}\, \ddot{d}_{\text{max}}\, T^2,
\end{equation}
where $\ddot{d}_{\text{max}} = 3\ \text{sign}(\Delta d)\ \text{m/s}^2$ and $\text{sign}(\Delta d) = +1$ if $\Delta d \geq 0$ and $-1$ if $\Delta d < 0$. The terminal lateral position, velocity, and acceleration of the quintic polynomial are then set as
\begin{equation}
\scalebox{0.94}{$
(d_f,\ \dot d_f,\ \ddot d_f) =
\begin{cases}
\left(
d_0 + \Delta d_{\max},\ 
\right. \\[4pt]
\left.
\dot d_{\max},\
\ddot d_{\max}
\right.)
& \text{if } |\Delta d_{\max}| < |\Delta d|, \\[8pt]
\left(d_{\text{off}},\ 0,\ 0\right)
& \text{otherwise},
\end{cases}
$}
\end{equation}
where $\dot d_{\max} = \dot d_0 + \ddot d_{\max} T$.
This ensures that the output paths are physically feasible, minimizing cumulative tracking errors typically introduced when adding a planner.
\definecolor{ourhighlight}{gray}{0.9}
\begin{table*}[t]
\centering
\caption{Driving performance and infraction analysis across benchmarks}
\label{tab:performance}
\resizebox{\linewidth}{!}{%
\begin{tabular}{l|l|cccc|ccccc}
\toprule
\multicolumn{1}{c|}{} & \multicolumn{1}{c|}{}
& Success & Driving & Route & Infrac. & Collision & Collision & Collision & Red light & Agent \\
\multicolumn{1}{c|}{Benchmark} & \multicolumn{1}{c|}{Method}
& rate & score & compl. & penalty & others & pedestrian & vehicle & infraction & blocked \\
& & (\%, $\uparrow$) & (\%, $\uparrow$) & (\%, $\uparrow$) & (\%, $\uparrow$)
& (\#/Km, $\downarrow$) & (\#/Km, $\downarrow$) & (\#/Km, $\downarrow$) & (\#/Km, $\downarrow$) & (\#/Km, $\downarrow$) \\
\midrule
\multirow{7}{*}{LB-all}
& Autopilot~\cite{roach}
& $61 \pm 14$ 
& $72 \pm 3$ 
& $90 \pm 4$ 
& $80 \pm 1$ 
& $\mathbf{0.00 \pm 0.00}$ 
& $\mathbf{0.00 \pm 0.00}$ 
& $\underline{0.23 \pm 0.10}$ 
& $0.54 \pm 0.03$ 
& $1.05 \pm 0.39$ \\
& Roach~\cite{roach} 
& $59 \pm 3$ 
& $75 \pm 2$ 
& $\underline{96 \pm 2}$ 
& $78 \pm 1$ 
& $\mathbf{0.00 \pm 0.00}$ 
& $0.04 \pm 0.04$ 
& $0.31 \pm 0.02$ 
& $0.12 \pm 0.01$ 
& $\underline{0.12 \pm 0.08}$ \\
& Roach + Rule~\cite{driveadapter} 
& $\underline{68 \pm 3}$ 
& $79 \pm 3$ 
& $95 \pm 1$ 
& $\underline{84 \pm 4}$ 
& $\mathbf{0.00 \pm 0.00}$ 
& $\mathbf{0.00 \pm 0.00}$ 
& $\mathbf{0.14 \pm 0.05}$ 
& $0.19 \pm 0.06$ 
& $0.15 \pm 0.06$ \\
& CaRL$^\star$~\cite{carl}  
& $39 \pm 3$ 
& $58 \pm 4$ 
& $73 \pm 1$ 
& $80 \pm 3$ 
& $\underline{0.22 \pm 0.04}$ 
& $\underline{0.01 \pm 0.01}$ 
& $0.27 \pm 0.04$ 
& $0.33 \pm 0.18$ 
& $1.44 \pm 0.43$ \\
\cline{2-11}
\rowcolor{ourhighlight} \cellcolor{white} & Ours w/o KF 
& $61 \pm 2$ 
& $79 \pm 2$ 
& $\underline{96 \pm 3}$ 
& $83 \pm 1$ 
& $\mathbf{0.00 \pm 0.00}$ 
& $\underline{0.01 \pm 0.01}$ 
& $0.34 \pm 0.04$ 
& $\mathbf{0.04 \pm 0.01}$ 
& $0.21 \pm 0.16$ \\
\rowcolor{ourhighlight} \cellcolor{white} & Ours 
& $67 \pm 3$ 
& $\underline{80 \pm 3}$ 
& $\mathbf{99 \pm 1}$ 
& $81 \pm 3$ 
& $\mathbf{0.00 \pm 0.00}$ 
& $\mathbf{0.00 \pm 0.01}$ 
& $0.31 \pm 0.08$ 
& $\underline{0.09 \pm 0.01}$ 
& $\mathbf{0.04 \pm 0.02}$ \\
\rowcolor{ourhighlight} \cellcolor{white} & Ours + Rule  
& $\mathbf{70 \pm 1}$ 
& $\mathbf{81 \pm 1}$ 
& $93 \pm 1$ 
& $\mathbf{87 \pm 1}$ 
& $\mathbf{0.00 \pm 0.00}$ 
& $\underline{0.01 \pm 0.01}$ 
& $\mathbf{0.14 \pm 0.03}$ 
& $0.18 \pm 0.03$ 
& $0.26 \pm 0.06$ \\
\midrule
\multirow{7}{*}{NCd-Town01}
& Autopilot~\cite{roach}
& $91 \pm 6$ 
& $81 \pm 2$ 
& $92 \pm 7$ 
& $84 \pm 1$ 
& $\mathbf{0.00 \pm 0.00}$ 
& $\mathbf{0.00 \pm 0.00}$ 
& $0.06 \pm 0.06$ 
& $1.29 \pm 0.58$ 
& $0.50 \pm 0.40$ \\
& Roach~\cite{roach}
& $73 \pm 6$ 
& $84 \pm 3$ 
& $93 \pm 2$ 
& $86 \pm 5$ 
& \underline{$0.05 \pm 0.08$} 
& \underline{$0.02 \pm 0.04$} 
& $0.38 \pm 0.23$ 
& $0.19 \pm 0.21$ 
& $0.21 \pm 0.23$ \\
& Roach + Rule~\cite{driveadapter}
& $88 \pm 4$ 
& $92 \pm 2$ 
& $92 \pm 4$ 
& $93 \pm 1$ 
& $\mathbf{0.00 \pm 0.00}$ 
& $\mathbf{0.00 \pm 0.00}$ 
& $\mathbf{0.04 \pm 0.01}$ 
& $0.29 \pm 0.06$ 
& $0.15 \pm 0.13$ \\
& CaRL$^\star$~\cite{carl}
& $48 \pm 0$ 
& $62 \pm 3$ 
& $84 \pm 4$ 
& $76 \pm 2$ 
& $0.45 \pm 0.34$ 
& $0.11 \pm 0.05$ 
& $0.28 \pm 0.07$ 
& $0.61 \pm 0.07$ 
& $1.13 \pm 0.88$ \\
\cline{2-11}
\rowcolor{ourhighlight} \cellcolor{white} & Ours w/o KF
& $85 \pm 5$ 
& \underline{$93 \pm 2$} 
& \underline{$99 \pm 2$} 
& $93 \pm 2$ 
& $\mathbf{0.00 \pm 0.00}$ 
& $\mathbf{0.00 \pm 0.00}$ 
& $0.22 \pm 0.09$ 
& $\mathbf{0.02 \pm 0.04}$ 
& \underline{$0.02 \pm 0.04$} \\
\rowcolor{ourhighlight} \cellcolor{white} & Ours
& \underline{$92 \pm 8$} 
& $\mathbf{95 \pm 3}$ 
& $\mathbf{100 \pm 0}$ 
& \underline{$95 \pm 3$} 
& $\mathbf{0.00 \pm 0.00}$ 
& $\mathbf{0.00 \pm 0.00}$ 
& $0.09 \pm 0.10$ 
& \underline{$0.12 \pm 0.08$} 
& $\mathbf{0.00 \pm 0.00}$ \\
\rowcolor{ourhighlight} \cellcolor{white} & Ours + Rule
& $\mathbf{95 \pm 6}$ 
& $\mathbf{95 \pm 6}$ 
& \underline{$99 \pm 2$} 
& $\mathbf{96 \pm 5}$ 
& $\mathbf{0.00 \pm 0.00}$ 
& $\mathbf{0.00 \pm 0.00}$ 
& \underline{$0.05 \pm 0.05$} 
& $0.13 \pm 0.16$ 
& $0.04 \pm 0.06$ \\
\midrule
\multirow{7}{*}{NCd-Town02}
& Autopilot~\cite{roach}
& $79 \pm 6$ 
& $67 \pm 5$ 
& $85 \pm 2$ 
& $74 \pm 6$ 
& $\mathbf{0.00 \pm 0.00}$ 
& $\mathbf{0.00 \pm 0.00}$ 
& $0.91 \pm 1.16$ 
& $2.70 \pm 1.01$ 
& $1.79 \pm 0.86$ \\
& Roach~\cite{roach}
& $65 \pm 5$ 
& $80 \pm 5$ 
& $85 \pm 10$ 
& $85 \pm 2$ 
& $\mathbf{0.00 \pm 0.00}$ 
& \underline{$0.03 \pm 0.05$} 
& $0.94 \pm 0.29$ 
& $0.34 \pm 0.32$ 
& \underline{$0.78 \pm 0.69$} \\
& Roach + Rule~\cite{driveadapter}
& $77 \pm 2$ 
& $82 \pm 0$ 
& $81 \pm 2$ 
& $88 \pm 2$ 
& $\mathbf{0.00 \pm 0.00}$ 
& $\mathbf{0.00 \pm 0.00}$ 
& \underline{$0.43 \pm 0.31$} 
& $0.72 \pm 0.25$ 
& $1.18 \pm 0.35$ \\
& CaRL$^\star$~\cite{carl}
& $15 \pm 2$ 
& $40 \pm 2$ 
& $53 \pm 2$ 
& $77 \pm 2$ 
& $\underline{2.52 \pm 1.36}$
& $0.04 \pm 0.06$ 
& $0.57 \pm 0.42$ 
& $2.24 \pm 0.50$ 
& $5.89 \pm 1.68$ \\
\cline{2-11}
\rowcolor{ourhighlight} \cellcolor{white} & Ours w/o KF
& $76 \pm 4$ 
& $87 \pm 2$ 
& $87 \pm 5$ 
& \underline{$93 \pm 3$} 
& $\mathbf{0.00 \pm 0.00}$ 
& $\mathbf{0.00 \pm 0.00}$ 
& $0.50 \pm 0.45$ 
& $0.17 \pm 0.07$ 
& $0.86 \pm 0.48$ \\
\rowcolor{ourhighlight} \cellcolor{white} & Ours
& \underline{$80 \pm 4$} 
& \underline{$88 \pm 2$} 
& $\mathbf{95 \pm 6}$ 
& $91 \pm 3$ 
& $\mathbf{0.00 \pm 0.00}$ 
& $\mathbf{0.00 \pm 0.00}$ 
& $0.58 \pm 0.27$ 
& $\mathbf{0.08 \pm 0.09}$ 
& $0.95 \pm 1.46$ \\
\rowcolor{ourhighlight} \cellcolor{white} & Ours + Rule
& $\mathbf{84 \pm 7}$ 
& $\mathbf{90 \pm 4}$ 
& \underline{$91 \pm 6$} 
& $\mathbf{95 \pm 2}$ 
& $\mathbf{0.00 \pm 0.00}$ 
& $\mathbf{0.00 \pm 0.00}$ 
& $\mathbf{0.36 \pm 0.21}$ 
& \underline{$0.13 \pm 0.11$} 
& $\mathbf{0.67 \pm 0.51}$ \\
\bottomrule
\end{tabular}}
\vspace{1mm}
\par
\footnotesize{LB-all denotes the average performance over 76 routes across Town01--Town06 in CARLA Offline Leaderboard v1. NCd denotes the dense setting in NoCrash, reporting results on Town01 and Town02, with 25 routes each. $^\star$: Results are obtained using the official checkpoint released in the authors' GitHub repository.  Boldface indicates the best performance, and underlined entries denote the second-best performance. Rule refers to the emergency brake rule proposed in DriveAdapter~\cite{driveadapter}. KF denotes the kinematic feasibility check.}
\vspace{-2mm}
\end{table*}

\section{EXPERIMENTS AND RESULTS}

\subsection{Benchmarks}

We evaluate the proposed method on the CARLA Offline Leaderboard v1~\cite{carla_leaderboard} and the NoCrash~\cite{nocrash} benchmark to validate its performance under diverse urban driving scenarios. All experiments are conducted using CARLA 0.9.10.1.

\textbf{CARLA Offline Leaderboard v1}~\cite{carla_leaderboard} provides a standardized evaluation protocol in complex urban environments, including dynamic traffic participants and traffic rules. Under this benchmark, we report the average performance over 76 predefined routes (Town01--Town06), with traffic density settings following~\cite{roach}.
It measures the ability of autonomous vehicles to complete navigation routes safely and efficiently under realistic traffic settings. Unlike the online Leaderboard, the offline benchmark strictly defines both the training and testing environments, enabling controlled and reproducible evaluation of generalization performance~\cite{roach}.

\textbf{NoCrash}~\cite{nocrash} is designed to evaluate the robustness of navigation policies under diverse environmental variations, including different weather conditions and maps. In this work, however, we adapt the protocol to analyze expert models that leverage privileged inputs and are trained on Town01–Town06. To eliminate unnecessary variations, we fix the weather to \textit{ClearSunset} and evaluate performance on Town01 and Town02 under the \textit{dense} traffic setting, with each town consisting of 25 predefined routes. The \textit{dense} traffic setting, characterized by heavy pedestrian and vehicle traffic, provides a challenging environment for collision avoidance and interaction handling. This setup avoids redundant experiments because our expert models leverage weather-invariant privileged inputs, allowing us to focus the evaluation on navigation performance under dense traffic.

\begin{table}[t]
\centering
\caption{Runtime analysis ($ms$) of different methods}
\vspace{-2mm}
\label{tab:runtime}
\begin{tabular}{l|ccc|c}
\toprule
Method & RL Inference & Planning & Control & Total \\
\midrule
Roach~\cite{roach}      & $\textbf{4.894}$ & - & - & $\textbf{4.894}$ \\
\rowcolor[gray]{0.9} Ours & $4.939$ & $1.100$ & $0.568$ & $6.607$ \\
\bottomrule
\end{tabular}
\vspace{2mm}
\par
\footnotesize{RL Inference denotes the network input–output processing time. Planning represents the time for trajectory generation, and Control the time required to produce control commands. Total denotes the overall inference time.}
\vspace{-4mm}
\end{table}

\subsection{Evaluation Metrics}

Performance is evaluated using the standard task-completion and safety-related metrics defined in the benchmark. All experiments are conducted with three different random seeds, and we report the mean and standard deviation to ensure statistical reliability. The definitions of the metrics are provided below (see~\cite{roach} for detailed descriptions).

\begin{itemize}

\item \textbf{Success rate} measures the percentage of routes successfully completed without terminal infractions, such as collisions with vehicles, pedestrians, or static obstacles.

\item \textbf{Driving score} reflects the overall navigation performance by combining route completion with penalties incurred from traffic violations.

\item \textbf{Route completion} indicates the percentage of the route completed, regardless of whether terminal infractions occur.

\item \textbf{Infraction penalty} represents the accumulated penalty factor applied due to traffic rule violations and unsafe behaviors.

\item \textbf{Collision others}, \textbf{Collision pedestrian}, and \textbf{Collision vehicle} measure the number of collisions per kilometer with static objects, pedestrians, and other vehicles, respectively.

\item \textbf{Red light infraction} measures the number of traffic light violations, normalized per kilometer.

\item \textbf{Agent blocked} measures the number of episodes terminated due to the vehicle being stuck or unable to proceed.

\end{itemize}

\subsection{Implementation Details}

Following the training protocol of our baseline, Roach’s RL experts~\cite{roach}, we adopt the same network architecture and identical RL hyperparameters, modifying only the output head to accommodate the proposed planning architecture. Training is performed on six CARLA maps (Town01--Town06) under randomized numbers of traffic agents and weather conditions. During both training and evaluation, the CARLA servers run at 10 FPS. The desired route is computed via A$^*$ search with start and goal points selected strictly following the protocol described in~\cite{roach}. All models are trained using a single NVIDIA RTX 3090 (24 GB) GPU. For fair comparison, all baseline and competing methods are reproduced under our experimental setup. Each algorithm is evaluated using the checkpoint that achieves the best performance within a total training budget of 10M steps. In addition, we use the parameter-tuned version of Autopilot reported in~\cite{roach}, which is optimized for improved performance.

\subsection{Main Results}

\textbf{Comparison with Autopilot}: As shown in Table~\ref{tab:performance}, our method achieves superior performance compared to Autopilot, a strong rule-based baseline on the CARLA Leaderboard v1 and NoCrash benchmarks. 
Without relying on the emergency brake rule, our approach improves the driving score by 8--21 points and the success rate by 1--6 points across the evaluated settings. 
In addition, our method significantly reduces infractions, achieving markedly improved results in all collision-related metrics (except collision vehicle), red-light infraction, and agent-blocked cases.
These results indicate that the proposed method can provide more effective action guidance for other end-to-end models. 
Furthermore, unlike Autopilot, our RL-based expert can supply informative latent features, enabling richer knowledge transfer.

\textbf{Comparison with Roach}:
Table~\ref{tab:performance} also shows that the proposed method achieves consistent improvements over Roach~\cite{roach}, a control-centric RL expert, with gains of 5--11 points in driving score and 8--19 points in success rate. 
In the NoCrash benchmark, both metrics approach nearly 100\%, suggesting that our method can generate substantially higher-quality ground truth compared to Roach. 
This improvement indicates that the proposed architecture effectively addresses the limitation of control-centric RL experts, which must implicitly internalize diverse and complex road geometries without a structured coordinate framework.
However, when Roach is combined with the emergency brake rule proposed in DriveAdapter~\cite{driveadapter}, it achieves a substantial performance improvement, even surpassing our method in terms of success rate on LB-all. 
In contrast, the performance gain from incorporating the same rule into our method is relatively marginal. 
This suggests that the proposed architecture already possesses sufficient braking capability without relying on heuristic rule-based emergency interventions. 
Furthermore, in terms of route completion, combining the rule with our method leads to degraded performance across all three benchmarks. 
We attribute this decline to overly conservative driving behavior induced by the heuristic emergency braking mechanism.

\textbf{Comparison with CaRL}:
As shown in Table~\ref{tab:performance}, the proposed method is further compared with CaRL~\cite{carl}, which simplifies the reward formulation of Roach~\cite{roach} while retaining the same baseline architecture. The proposed method consistently outperforms CaRL in terms of both driving score and success rate. For the evaluation of CaRL~\cite{carl}, we use the checkpoint officially released by the authors in their GitHub repository. However, a fully fair comparison cannot be guaranteed because the released checkpoint was trained under vehicle dynamics and simulation settings different from those used in our experiments, including a different ego vehicle model and a different CARLA version.

\begin{figure}[!t]
\centering
\includegraphics[width=1.0\columnwidth]{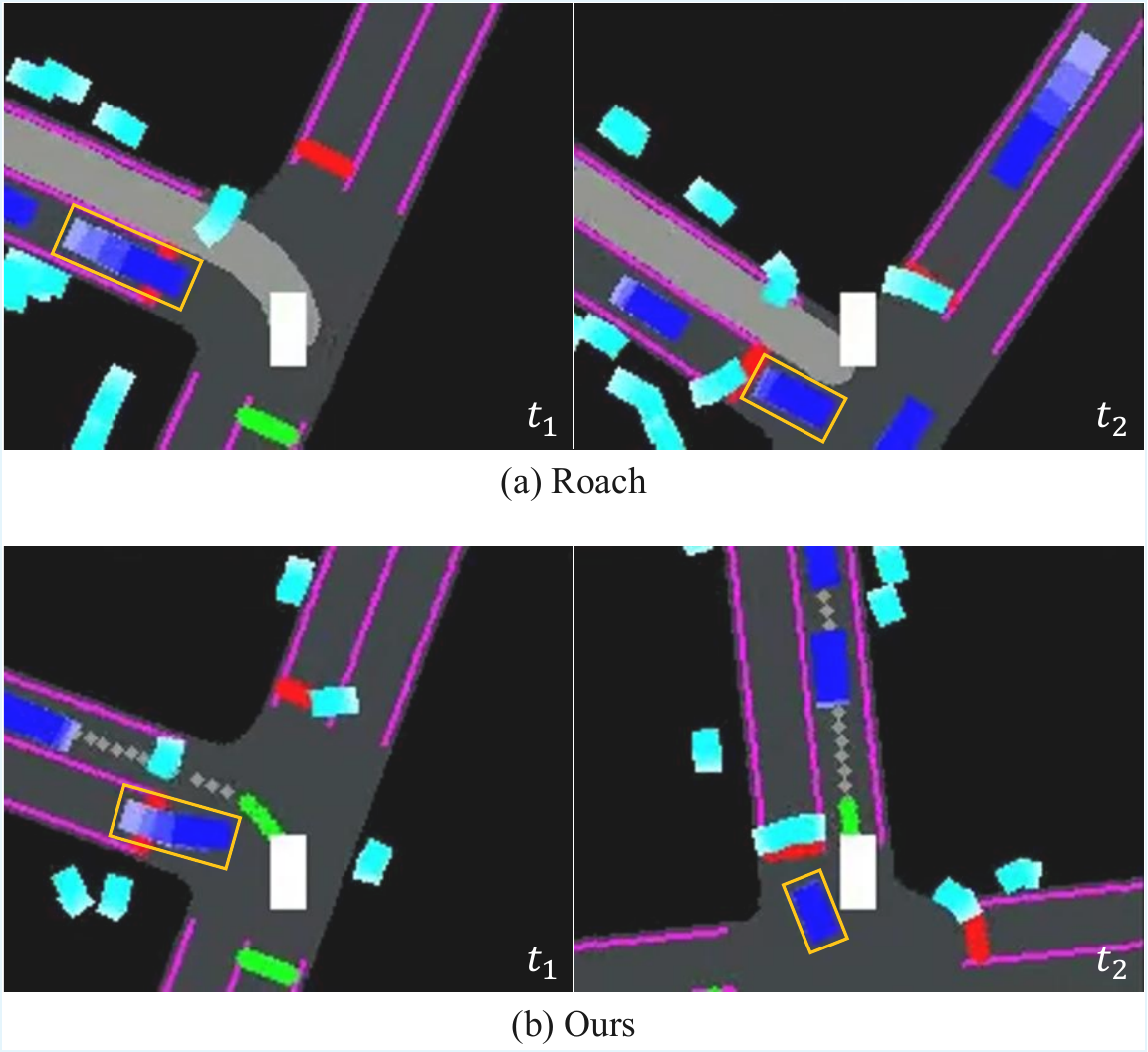}
\caption{Visualization of driving scenarios in NCd-Town02 for (a) Roach and (b) Ours at two time steps, 
$t_1$ and the subsequent time step $t_2$. Pedestrians and vehicles are shown in cyan and blue, respectively, and are visualized with their historical states up to $-1.5$, $-1.0$, and $-0.5$ seconds. The ego vehicle is shown in white, the route in light gray, the road in dark gray, and road boundaries in magenta. The color shown at the end of the road represents the traffic light state. Yellow bounding boxes indicate vehicles approaching the ego vehicle and leading to a collision.}
\vspace{-12pt}
\label{qualitative}
\end{figure}

\subsection{Ablation Study}

\textbf{Effect of the Kinematic Feasibility Check}: An ablation study is conducted to evaluate the effectiveness of the proposed kinematic feasibility check (KF). As shown in the gray-highlighted entries of Table~\ref{tab:performance}, our method consistently outperforms the variant without KF (Ours w/o KF) in terms of success rate and driving score across all benchmarks (LB-all, NCd-Town01, and NCd-Town02). These results demonstrate that incorporating the kinematic feasibility check into the trajectory planning process enables the RL policy to generate kinematically trackable trajectories, thereby minimizing cumulative tracking errors that may arise when an additional planner is introduced. Ultimately, this leads to improved overall driving performance.

\textbf{Runtime Analysis}: We investigate how the runtime changes compared to Roach~\cite{roach} when incorporating the proposed RL-based trajectory planning architecture, which introduces additional Planning and Control modules. The runtime reported in Table~\ref{tab:runtime} is measured as the average over NCd-Town01. The results show that our method incurs an additional latency of approximately 1.713\,ms compared to Roach due to the added Planning and Control stages. Nevertheless, this corresponds to an inference speed of around 151\,FPS, which remains substantially faster than the 20--30\,FPS typically required for real-world autonomous driving decision-making systems.

\subsection{Qualitative Analysis}

We include Fig.~\ref{qualitative} to qualitatively compare the proposed method with Roach~\cite{roach}. First, the green trajectory attached to the ego vehicle in Fig.~\ref{qualitative} illustrates the intended target direction of the proposed method, unlike Roach, thereby significantly improving the interpretability of the algorithm. Moreover, this explicit trajectory enables trajectory-level guidance for end-to-end models supervised by the proposed expert. The visualization results further show that both Roach and Ours successfully avoid the approaching vehicle that leads to a collision (yellow bounding box). However, after the avoidance maneuver, the proposed method successfully returns to the route, whereas Roach fails to rejoin the route and becomes stuck at that location. This demonstrates that the formulation of the proposed planning architecture is more effective than Roach on roads with large curvature.
\section{CONCLUSION AND FUTURE WORK}

This paper presents a trajectory planning architecture for RL-based driving experts that addresses the inherent limitations of control-centric approaches. 
By operating in the Frenet-frame coordinate system and explicitly incorporating kinematic feasibility into trajectory generation, the method simplifies the representation of complex road geometries while mitigating cumulative tracking errors introduced by an additional planner.
Unlike conventional RL experts that output direct control commands, the proposed framework provides structured trajectory-based guidance, thereby improving interpretability and aligning expert supervision with modern end-to-end planning architectures. 
Extensive evaluations on the CARLA Offline Leaderboard v1 and NoCrash benchmarks demonstrate that the proposed approach consistently outperforms state-of-the-art control-based RL experts in terms of driving score and success rate. 
In future work, we plan to further validate the proposed architecture on more recent and challenging benchmarks, such as Bench2Drive~\cite{bench2drive} and Longest6~\cite{transfuser}, to assess its generalization capability. In addition, we will investigate how effectively the proposed trajectory-centric expert can supervise end-to-end autonomous driving models---particularly those built upon planning architectures---and quantify the performance gains achieved over conventional control-based expert supervision.

\addtolength{\textheight}{0cm}

\bibliographystyle{IEEEtran}
\bibliography{Section/Ref}

\end{document}